# Predict joint angle of body parts based on sequence pattern recognition

Amin Ahmadi Kasani
*Department of Mathematics, Statistics and Computer Science, College of Science, University of Tehran*
Tehran, Iran
aakasani@ut.ac.ir

Hedieh Sajedi
*Department of Mathematics, Statistics and Computer Science, College of Science, University of Tehran*
Tehran, Iran
hhsajedi@ut.ac.ir

***Abstract*—The way organs are positioned and moved in the workplace can cause pain and physical harm. Therefore, ergonomists use ergonomic risk assessments based on visual observation of the workplace, or review pictures and videos taken in the workplace. Sometimes the workers in the photos are not in perfect condition. Some parts of the workers' bodies may not be in the camera's field of view, could be obscured by objects, or by self-occlusion, this is the main problem in 2D human posture recognition. It is difficult to predict the position of body parts when they are not visible in the image, and geometric mathematical methods are not entirely suitable for this purpose. Therefore, we created a dataset with artificial images of a 3D human model, specifically for painful postures, and real human photos from different viewpoints. Each image we captured was based on a predefined joint angle for each 3D model or human model. We created various images, including images where some body parts are not visible. Nevertheless, the joint angle is estimated beforehand, so we could study the case by converting the input images into the sequence of joint connections between predefined body parts and extracting the desired joint angle with a convolutional neural network. In the end, we obtained root mean square error (RMSE) of 12.89 and mean absolute error (MAE) of 4.7 on the test dataset.**

## I. INTRODUCTION

Is it possible to guess the location of organs in the body that are not visible in a two-dimensional (2D) image, and measure the angle of their joints based on the other body positions? This was the question we were trying to answer.

In cases where ergonomists are trying to measure the hazards of the work environment, they use snapshots of the environment to calculate body risk assessment, such as the Rapid Upper Limb Assessment (RULA) [4] score described in Section I.A. We chose RULA because it is commonly used in research papers related to the ergonomic field[1]. Collecting required data for desired calculations can be done well with good enough accuracy by using three-dimensional(3D) data or human observation, but 2D images have little information about the depth of field. So, in computer vision processing, high-error photos that do not have information about some joint and bone positions may be left out because they do not have enough data for ergonomic assessments.

With the OpenPose[2] tool described in Section I.B., the body joints that can be seen in the 2D image can be well identified in most cases. We choose OpenPose because, in new ergonomics-related research, it performed better than Kinect-based system[3], which is 3D based system body recognition. We will use this tool to convert input images to a 2D sequence matrix of body parts detected from the input. After that, we will create a 3D matrix with a relationship between each key point like Figure 2. Then we can use this relationship to identify the pattern of sequence and predict the desired angle with a convolutional neural network. We used Convolutional Neural Network (CNN) because it is well known for pattern recognition, and in our investigation, it performed way better than Recurrent Neural Network (RNN) methods.

Therefore, if we create a model that can analyze detected body parts and predict angles based on something like human experience, it can probably provide a good estimate of the location of body parts and predict that joint angle we can not see directly in this paper, we try to test this theory.

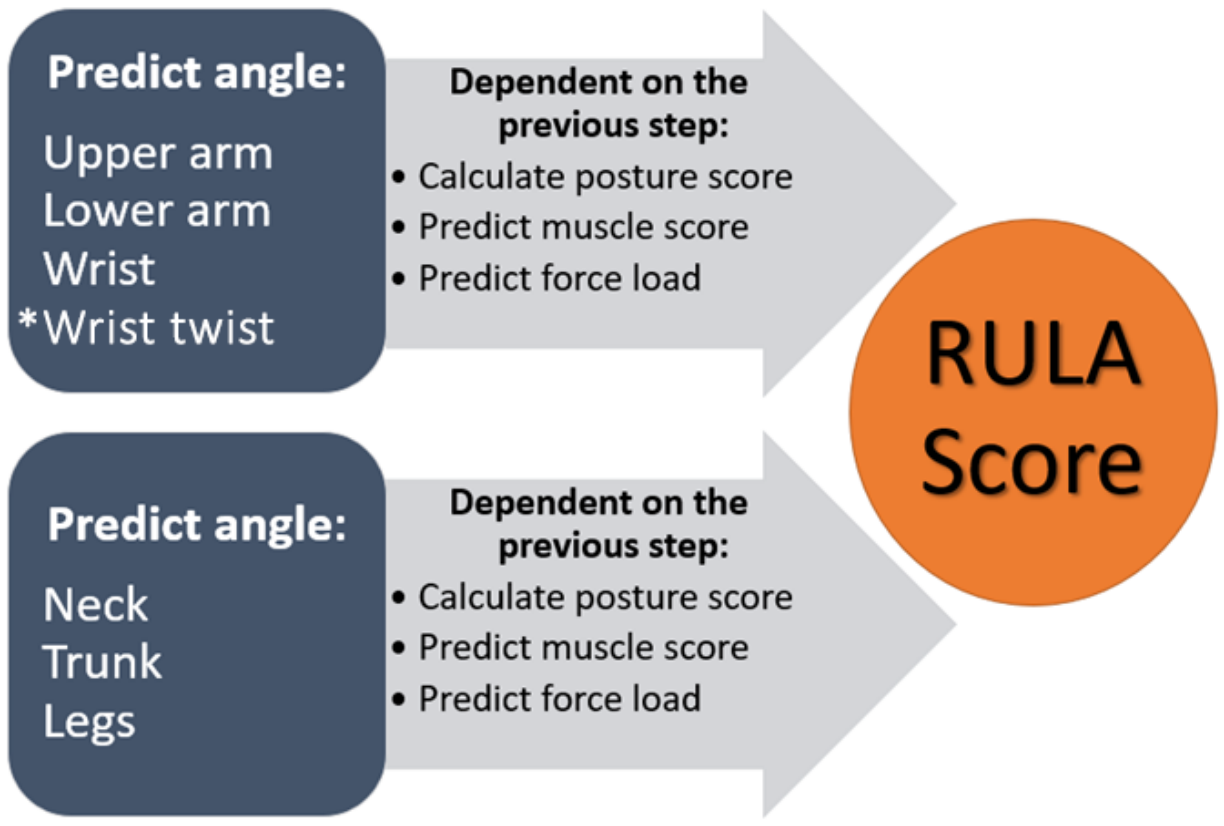

Fig. 1. Explains the process of calculating the RULA score [4]. We are trying to examine the results of the first part. * except for the wrist part, which is usually dependent on the tools used by the worker.

### *A. RULA*

The RULA [4] is designed to evaluate the exposure of individual workers to ergonomic risk factors associated with upper extremity Musculoskeletal Disorders (MSD). The ergonomic assessments are necessary because they cause detection and prevention of Work-related Musculoskeletal Disorders (WMSDs), which are ranked second worldwide in shortening people working [5]. Thus, the position of each part of the body is calculated separately in four parts presented in Figure1. The first two of them are related to the angles of body joint's and the other two parts are the muscles score and the force score, the final score being the amount of pressure on the worker's body by a numerical value. In this paper, we want to measure the group's angles required in the first part.

The RULA score is usually obtained by direct human observation or from digital snapshots like photos and videos. Ergonomists typically try to find the most critical posture or perform in long-duration by workers to predict reliable RULA scores. In these cases, using video is usually better than using photos, but this does not affect the solution we provide.

### B. *OpenPose*

Recognizing the location of body parts in 2D images played a crucial role in our research. An open-pose 25-body model receives an image as input and outputs the location of each person's body joints along with a number indicating the accuracy of the locations. The model gives us a 3D matrix for each person detected from the input image we have a layer. We have 25 rows and three columns filled with each body part position and accuracy in each layer. We use this output to build the structure we need for our joint angle predictor input described in section III.

OpenPose[2] is an open-source project developed by researchers at Carnegie Mellon University written in C++ programming language and Caffe deep learning framework. We use OpenPose by using its Python Application Programming Interface (API) in preprocessing stage of our final model. The output of the 25-body model that we are using here is presented as key points in Figure 2.

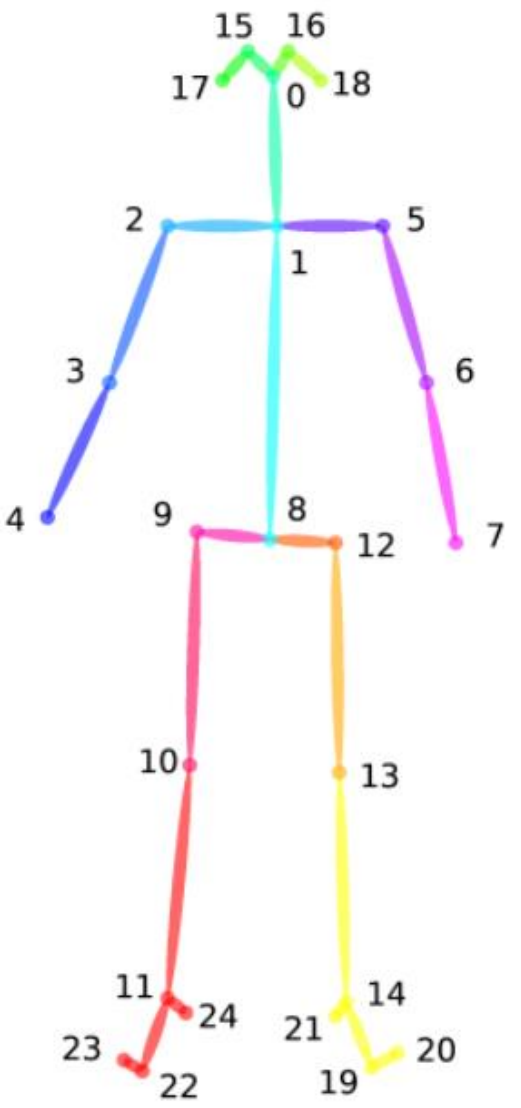

Fig. 2. The output of the OpenPose for 25 skeleton and body joints model is the key points named 0 to 24.

### C. *Joint angle estimation*

To calculate the RULA score, we need to obtain the angles of the various joints of the person's body. To do this, we prepared a specific dataset with an accurate enough joint angle degree tag to create a neural network to predict the angles of the joints we wanted in output. We also created relationships between key points generated by the OpenPose 25-body model instead of using only key points location to improve the accuracy of measuring angles. Finally, a neural network is needed to process the relationships between key points and predict angles from different viewpoints. More information is available in the Methods section and Figure 4.

## II. Related Work

Our work is inspired by (Ergonomic risk assessment based on computer vision and machine learning[5]), they tested the possibility of calculating the RULA score using 2D visual input and OpenPose. They created new key points based on detected body key points to improve performance in situations where some of the body parts are not visible besides that they tried to fix some of the undetected body parts. They used the accuracy confidence parameter of OpenPose to extract noisy images, then with the help of seven ergonomists they evaluated the feasibility and result of this method. At the end of the analysis, they confirmed the feasibility of this method. They used involved joints from Table I was used to identify the required angles to calculate the RULA score. They did not publish comparable results with our results but compared to seven ergonomist calculations, they achieved 0.6 Cohen's in their real-world experiment.

Another approach to predict body joint angle by processing Electromyography (EMG) signals require special tools pre-installed on the body. This makes all the information of the body posture available for processing at any time, and even self-occlusion does not occur, which in turn increases the final accuracy. In [6], they reached Root Mean Square Error (RMSE) less than 3.5 degrees, and in [7], they earned RMSE 2 degrees. However, this method is not applicable in RULA score calculation applications.

The research of [8] introduced a new method for dynamic hand gesture recognition with leap motion. We tried to use a similar approach with Long Short-Term Memory(LSTM)[9] on OpenPose model output and extract the desired joint angle, but the results were unsatisfactory. We will discuss the problem in Section III.

TABLE I. Connection between RULA joint angles and OpenPose 25 skeleton data.

| Angle name | Acronym | Involved joints |
|---|---|---|
| Left elbow | EL | ∠04,03,02 |
| Right elbow | ER | ∠05,06,07 |
| Left Shoulder | SL | ∠03,02,09 |
| Right Shoulder | SR | ∠06,05,12 |
| Left Shoulder 2 | SL2 | ∠03,02,01 |
| Right Shoulder 2 | SR2 | ∠06,05,01 |
| Left Knee | KL | ∠09,10,11 |
| Right Knee | KR | ∠12,13,14 |
| Neck twisting | NT | *∠00,01,02 |
| Neck bending left | NBL | ∠17,01,02 |
| Neck bending right | NBR | ∠18,01,05 |
| Neck flexion | NF | ∠00,01,08 |
| Trunk twisting right | TTR | ∠02,08,09 |
| Trunk twisting left | TTL | ∠05,08,12 |
| Trunk Bending | TB | ∠09,08,01 |
| Trunk flexion | TF | *∠(10 and 13),08,01 |

*They replaced 00 with middle ear and used 10 and 13 joint to create middle knee point.

According to work done in the past, if the values of the angles can be obtained on an image, the variations of the angles can also be obtained at intervals from the worker's videos. Furthermore, we can calculate more information about body parts movement, such as variance, to estimate the RULA score with even better accuracy and reliability, which shows the possibility of using this method in more complex tasks. The contribution of our work is investigating the problem in the circumstances that the 2D images or viewpoint of worker's positions are not ideal or body parts are noisily detected.

## III. METHODS

In the method of this paper, we tried to keep noisy images with low accuracy in OpenPose model output that did not detect all the major joints of the input image, for example, in the image of Figure 3. In some situations, if a worker has a box in his hand and his right elbow is 90 degrees, his left hand is probably at a 90-degree angle. We tried to use all the body key points detected by OpenPose to predict each joint angle with the same approach. We wanted to check the non-ideal conditions, so we chose the following settings for OpenPose. We set the net_resolution parameter on "-1x288" which is lower than the default value "-1x384" to reduce workload and resource consumption on the computer. For example, check the two experimental images in Figure 3, there is no information detected from the left hand.

To test this approach, the angles obtained by the ergonomists were not enough because they can not see the left hand too. So we created a new real-world dataset in 185 pre-selected postures checked each joint angle with a protractor in 1201 images from different viewpoints combined with 1018 images from 121 postures on a 3D artificial human model to perform harder and painful postures to train the neural network. For test data, we selected 33 postures not necessarily performed in the train dataset and created a real-world test set with 233 images from different viewpoints, dataset is accessible at [10]. As preprocessing part, we created a data extractor tool to run each image on OpenPose 25-body and skeleton model with "number_people_max"=1 parameter because the final model should work with only one person data as input. We store results for our final model training. Then we used body key points to create a relation between key points like points 2 and 3 in Figure 2.

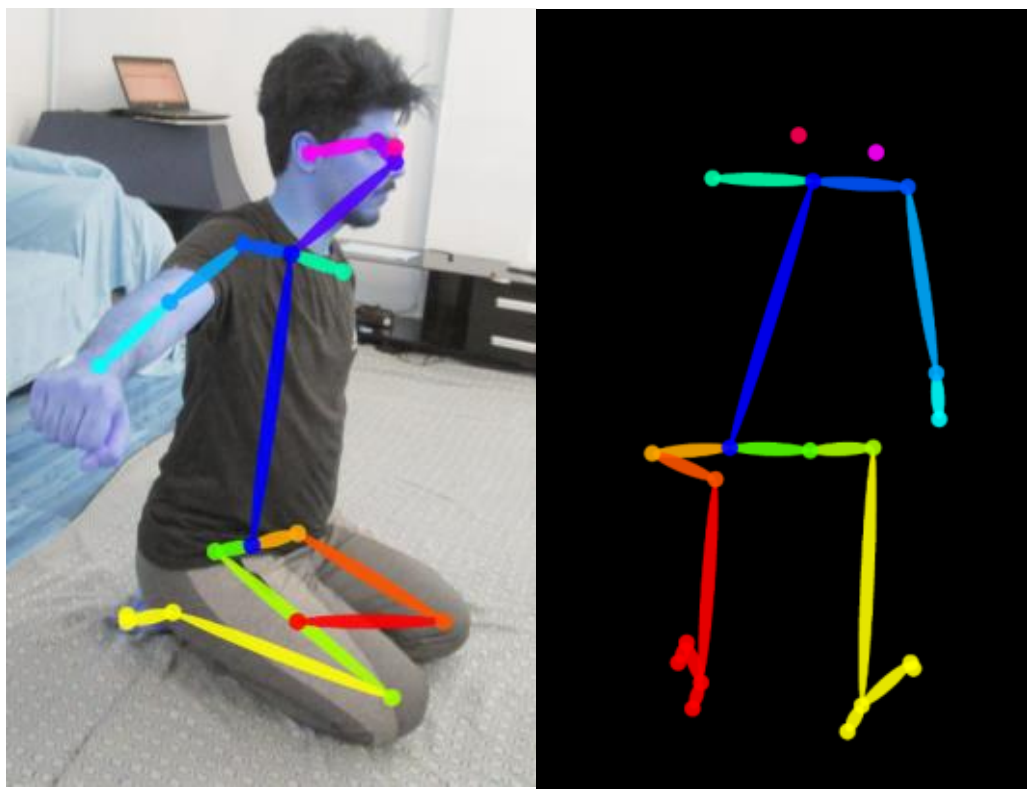

Fig. 3. Examples of self-occlusion that is responsible for undetected body parts.

The OpenPose model is not sensitive to different environments and different cameras, our dataset images were taken by a regular canon A3400 camera at 8MP resolution, and 3d models are 1MP. In preprocessing part, we can remap all detected parts to close similar size, and after that, we normalize images location by scaling every detected body part location with 1/6000 to 0-1 range. In addition, the person inside the image could be in different parts of the image. Hence, we remap person detected body part location from the input image with key point 8 in Figure 2 to the center of arbitrary space ([0.5,0.5]) that key point location could be in. With this technique, we improved the final model with the similarity between key points location in model input to detect the relationship between body parts. We decreased the sensitivity of person location from the input.

The images in the dataset are recorded at different distances by using the Data Augmentation method and multiplying the location of points obtained from the OpenPose model; we reduced the effect of varying the camera's distance from the person and increase sensitivity on the relationship between parts, not sizes. The final part of preprocessing was converting key points data into a more meaningful shape. Instead of letting the neural network itself figure out these relationships, we created a joint connector function and connected the following key points. We retained the previous connections in Figure 2 and added new connections that can be seen in Figure 4.

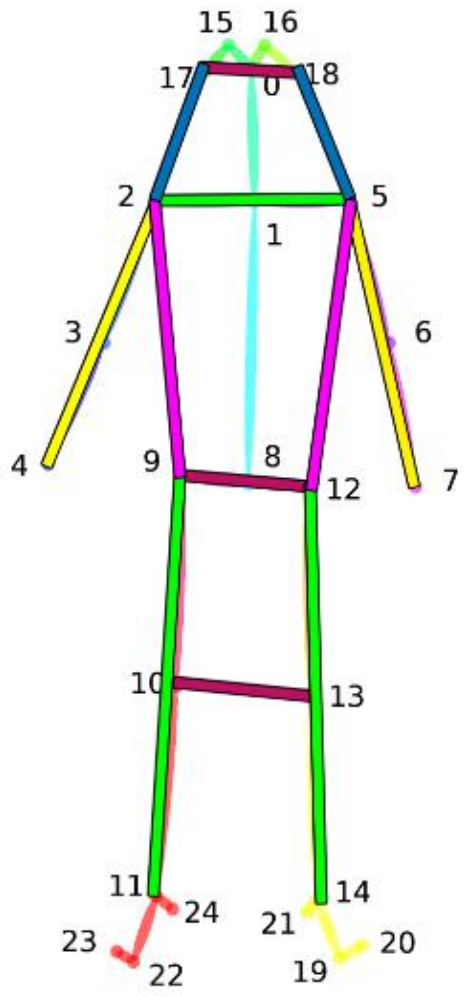

Fig. 4. Shows the new relationships between key points created by the joint connector function as colored lines.

The initial solution was to try to build a big enough neural network to predict 16 joint angle outputs we need from person body point positions. Still, because of the diversity inside body points that the OpenPose model does not detect, it was impossible to build such a neural network with both accurate and acceptable speed. So the better solution was to create 16 small neural networks to identify each of the 16 desired joint angles that we need for the RULA score. After checking multiple neural network architecture with learning rate 0.0001, Batch size 128, optimizer Adam[11] and loss RMSE and metrics Mean Absolute Error[12](MAE), and Mean Squared Error (MSE), we created a TensorFlow [13] model with a simple and stacked Bidirectional LSTM and normal LSTM with 256 units with Rectified Linear Unit (ReLU) activation function followed by a 50 unit FF layer with ReLU activation and a final dense layer as output. The model was small but not accurate enough compared to the final architectures we achieved.

As we have said before, we will examine the issue in this section briefly. Usually, RNN models are sensitive to missing data. Lots of research has been done in this area [14], but we wanted to create 16 small models. The input that we have has many missing parts because of undetected body parts by the OpenPose model. We found another simple way to extract desired data from the input sequence. We started with the Masking layer to reduce the effect of missing data with 0 attributes, followed by 3, 2D convolution layer separated by dense layer in between and MaxPooling layer before the last convolution layer with a small dense layer as output. We reduced the error of the RNN method to 1/3 with the CNN model. Finally, a program is created that loads all 16 models we made before for each joint angle to extracts all 16 joint angles RULA score need from the input matrix containing person body part positions, preprocessing, model tester source code and final modes are available at[10]. In Figure 5, we explained the preprocessing steps in summary.

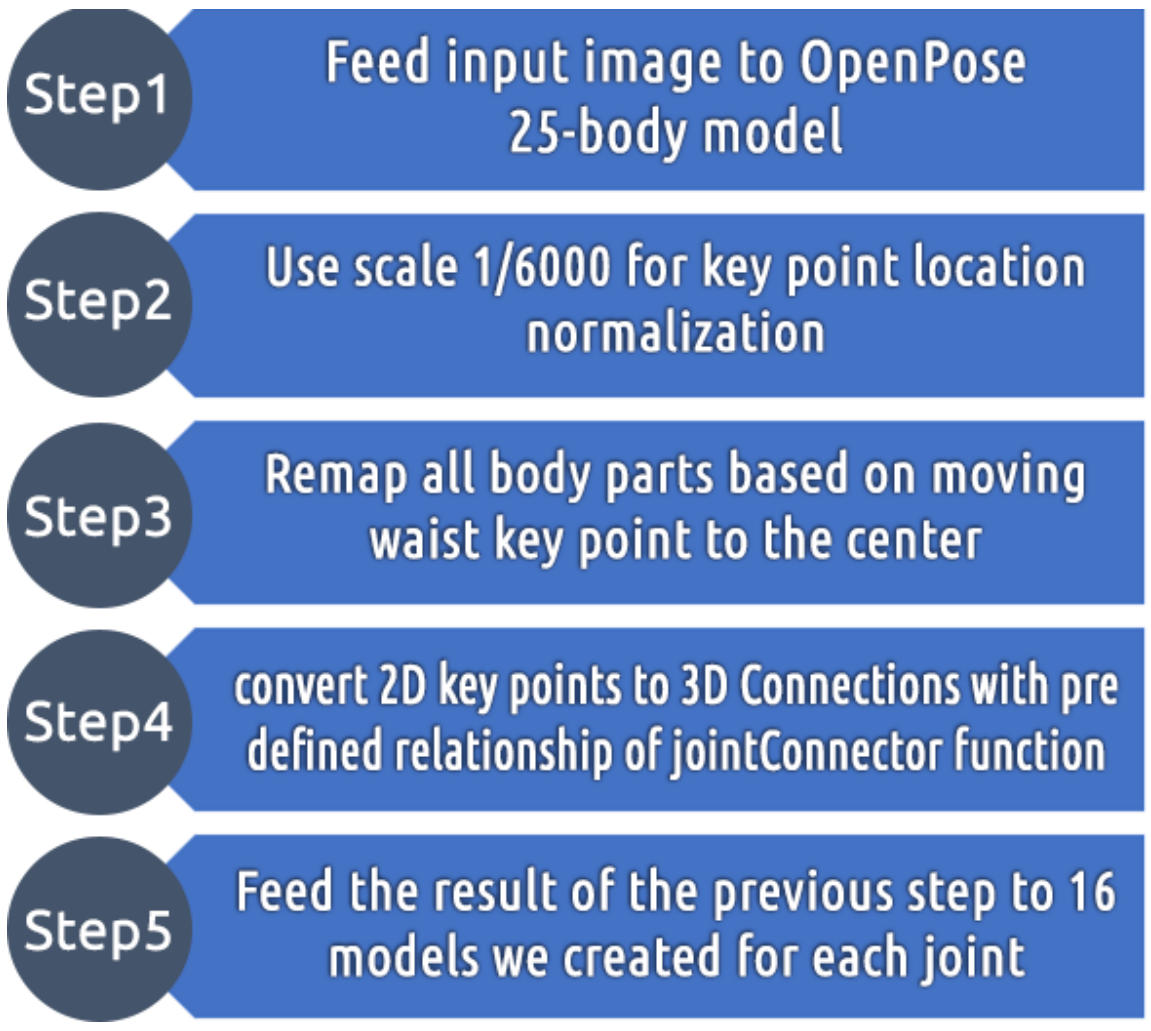

Fig. 5. Explain 5 step process of measuring joint angles from a new image.

## IV. RESULTS

The RULA score in most cases is not sensitive to small errors even in some cases ±25 degree error in angle detection, does not change the final score, so by choosing RMSE loss we tried to focus more on reducing the big error part between actual angle and predicted angle. We got the following angle prediction error results in Table II from each dependent model:

TABLE II. The final results of the training set and test set were presented for each angle separately.

| Angle name | Train Error(degree) | Test Error(degree) |
|---|---|---|
| Left elbow | 4.11 | 4.96 |
| Right elbow | 8.11 | 6.27 |
| Left Shoulder | 9.31 | 6.5 |
| Right Shoulder | 11.68 | 10.63 |
| Left Shoulder 2 | 2.81 | 7.07 |
| Right Shoulder 2 | 7.81 | 11.05 |
| Left Knee | 5.44 | 7.69 |
| Right Knee | 5.20 | 7.28 |
| Neck twisting | 1.82 | 2.30 |
| Neck bending left | 0.98 | 0.90 |
| Neck bending right | 1.60 | 1.80 |
| Neck flexion | 1.16 | 1.81 |
| Trunk twisting right | 0.52 | 1.03 |
| Trunk twisting left | 0.68 | 1.28 |
| Trunk Bending | 1.44 | 1.04 |
| Trunk flexion | 0.78 | 3.44 |

According to the results of Table III, it can be seen that the accuracy of predicting the angle of the hands is less than other angles. However, the final model predicts the proximity of the actual angle of the hands. But due to the availability of cheaper solutions, like inertial measurement sensor units in the future[15], we can combine these two methods to achieve high accuracy of prediction in adverse visual conditions.

TABLE III. The final results of the training set and test set are presented for average all angles.

| | RMSE Error | MAE Error |
|---|---|---|
| Train Data | 9.6150 | 3.7383 |
| Test Data | 12.8905 | 4.7735 |

In the end, we selected a random picture from the test set in which some of the body parts are not in viewpoint and check the accuracy of that body part joint angle. The output of OpenPose can be seen in Figure 6. The right elbow should be 0 degrees and predicted 0 degrees, the right shoulder should be 45 degrees and predicted 41.40 degrees, and the right shoulder-2 should be 0 degrees and predicted 0, which is accurate enough.

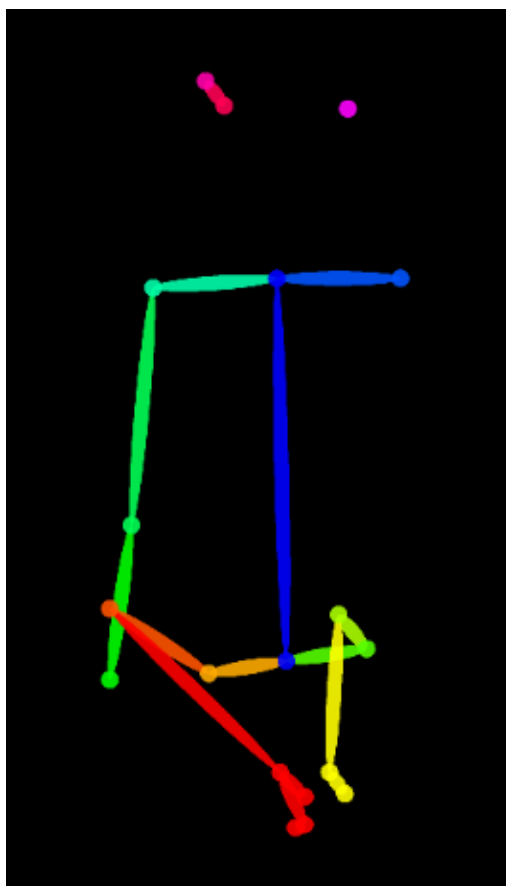

Fig. 6. A randomly selected image from the test set with undetected right-hand key points.

## V. CONCLUSION

In this paper, we examined the possibility of predicting the angle of body parts from the sequence of key points detected by the OpenPose tool with moderate accuracy. We created a new database with a variety of positions and viewpoints, then found a solution to connect the detected points and improved the final accuracy in this way, examined a solution to make the output of the OpenPose model closer to each other in terms of location in similar positions. Eventually we examined two neural network structures for the final model to process the sequence. We found the strengths and weaknesses of this method, and to solve them, we proposed using sensor bracelets with two-dimensional image processing. In the end, we reached RMSE 12.89 and MAE 4.7 on the test dataset, and randomly selected an image and compared the angle of the undetected part with its actual value, and the result was acceptable.